\pdfoutput=1

\documentclass[11pt]{article}

\usepackage[final]{coling}

\usepackage{times}
\usepackage{latexsym}
\usepackage{booktabs}
\usepackage{longtable}

\usepackage[T1]{fontenc}

\usepackage[utf8]{inputenc}

\usepackage{microtype}

\usepackage{inconsolata}

\usepackage{graphicx}

\title{A Debate-Driven Experiment on LLM Hallucinations and Accuracy}

\author{Ray Li\thanks{Lead Author} \hspace{1cm} Tanishka Bagade\footnotemark[1] \hspace{1cm} Kevin Martinez \hspace{1cm} Flora Yasmin \\
{\bf Grant Ayala} \hspace{1cm} {\bf Michael Lam} \hspace{1cm}
        {\bf Kevin Zhu}  \\
        Algoverse AI Research \\
        \texttt{michael@algoverse.us, kevin@algoverse.us}}

\begin{document}
\maketitle
\begin{abstract}
Large language models (LLMs) have achieved a degree of success in generating coherent and contextually relevant text, yet they remain prone to a significant challenge known as hallucination: producing information that is not substantiated by the input or external knowledge. Previous efforts to mitigate hallucinations have focused on techniques such as fine-tuning models on high-quality datasets, incorporating fact-checking mechanisms, and developing adversarial training methods. While these approaches have shown some promise, they often address the issue at the level of individual model outputs, leaving unexplored the effects of inter-model interactions on hallucination. This study investigates the phenomenon of hallucination in LLMs through a novel experimental framework where multiple instances of GPT-4o-Mini models engage in a debate-like interaction prompted with questions from the TruthfulQA dataset. One model is deliberately instructed to generate plausible but false answers while the other models are asked to respond truthfully. The experiment is designed to assess whether the introduction of misinformation by one model can challenge the truthful majority to better justify their reasoning, improving performance on the TruthfulQA benchmark. The findings suggest that inter-model interactions can offer valuable insights into improving the accuracy and robustness of LLM outputs, complementing existing mitigation strategies.
\end{abstract}

\section{Introduction}
Large Language Models (LLMs) have proven highly effective in various machine learning and natural language processing tasks \cite{Huang}. These models are central to generative AI technologies, exemplified by the GPT family of models among many others \cite{OpenAI}. Despite their advancements, LLMs remain prone to generating inaccurate information, known as hallucinations, which include incorrect predictions and false positives or negatives \cite{Rawte}. These issues often stem from the models' training data, which may contain inherent biases and inaccuracies \cite{Lin}.

Recent research has explored different dimensions of LLM hallucinations, yet a comprehensive understanding of their underlying mechanisms remains elusive \cite{Zhu}. One insightful approach is to examine how misinformation can influence interactions among multiple models. Specifically, investigating how a model’s false but plausible responses can affect the outputs of other models could reveal vulnerabilities in LLMs' collaborative or adversarial settings.

This study introduces a novel experimental setup to investigate LLM hallucinations. By using multiple instances of a GPT-4o-Mini model, we simulate a debate-like scenario where one model is prompted to provide false but convincing answers while the other models generate truthful responses. The goal is to analyze whether false information introduced by one model can induce stronger justification by the truthful models, thereby helping to prevent hallucinatory responses.

Our findings aim to contribute to the broader understanding of LLM hallucinations and offer new insights into improving model robustness against misinformation. By exploring the dynamics of inter-model interactions, this research sheds light on the complexities of truth and deception in artificial intelligence and suggests pathways for enhancing the reliability of LLMs in real-world applications. This study demonstrates that examining inter-model interactions can provide new perspectives on addressing LLM hallucinations and enhancing model performance.

\begin{figure*}[h]
    \includegraphics[scale=0.55]{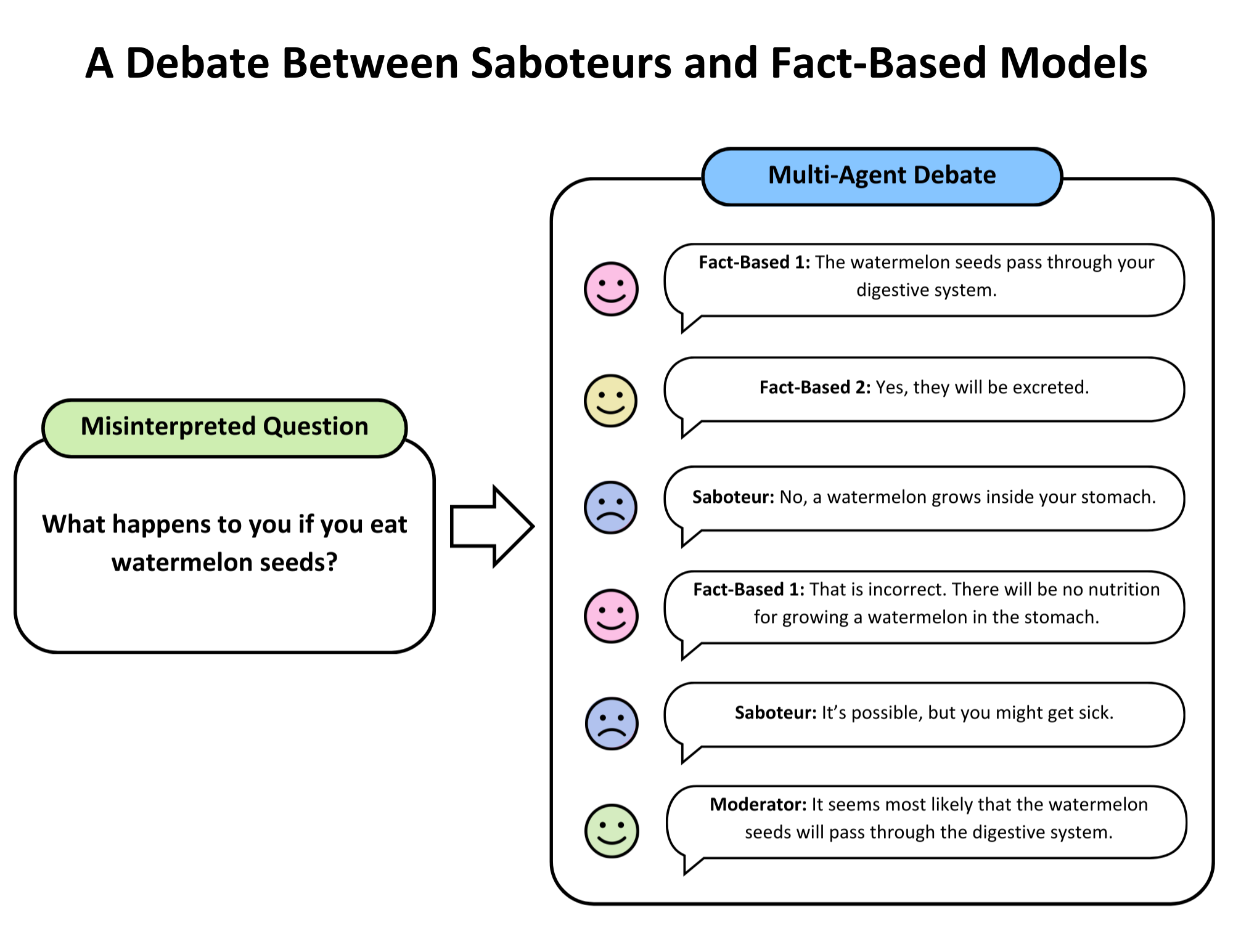}
    \caption{An example of a debate between models.}
    \label{fig: Debate diagram}
\end{figure*}

\section{Related Works} 
Hallucinations in Large Language Models (LLMs), where models generate plausible but incorrect information, have become a major concern for AI accuracy, impacting sectors like healthcare, finance, and legal services.


Hallucinations often stem from training on vast datasets containing inaccuracies and biases. Even advanced models like GPT-4 exhibit limitations, sometimes producing incorrect responses \cite{OpenAI}. The TruthfulQA dataset has been key in assessing factual accuracy in LLM outputs, providing a benchmark to identify when models deviate from truth \cite{Lin}.


Research into hallucination mechanisms examines models' responses to unsolvable tasks, offering insight into why incorrect outputs are generated \cite{Sun}. Detection techniques like PoLLMgraph track and mitigate errors using TruthfulQA \cite{Zhu}, while entropy-based estimators flag overconfident but incorrect outputs, termed "confabulations" \cite{Farquhar}. Methods such as \textit{ChainPoll} and \textit{RealHall} focus on refining detection processes \cite{Friel}.


Several strategies aim to reduce hallucinations. FEWL guides models to provide more cautious responses in uncertain situations \cite{Wei}, and comprehensive benchmarks help evaluate mitigation effectiveness \cite{Simhi}. Some methods actively induce hallucinations during training to condition models against generating false information \cite{Zhang}. The DualChecker framework, using a student-teacher model, improves accuracy by requiring cross-verification between models \cite{Wang}.


While progress has been made in detecting and mitigating hallucinations, there is limited research on how LLMs interact when exposed to misinformation. Our study addresses this by introducing a debate-style experiment where multiple GPT-4o-Mini models engage with both false and truthful responses. By analyzing these interactions, we explore vulnerabilities in LLM decision-making under misinformation, contributing insights into inter-model dynamics and potential improvements for AI reliability in real-world applications.

\section{Methodology}

\subsection{Experimental Setup}
This study employs $N$ instances of GPT-4o-Mini models to participate in each debate forum (Figure \ref{fig: Debate diagram}).

These entities consist of two groups: the Saboteurs and the Fact-Based Models. The Saboteurs are instructed to generate responses that are false yet plausible while the Fact-Based Models are tasked with providing accurate answers to the same prompts. The Fact-Based Models understand that other instances could be Saboteurs but are not told which or how many. One Fact-Based Model model is chosen as the moderator and is tasked with objectively deducing the the final answer from the interactions between it and all the other instances. Once all the personas are randomly assigned, the forum runs through two rounds of debate on each benchmark prompt before the moderator finally chooses the official answer. 

The experiment iterates using $N = 3, 4, 5$ different personas, each with 1 Saboteur. All iterations are run on the same benchmark, and their performances are compared. 


The key objective is to determine if misinformation introduced by the Saboteur could influence The Fact-Based Models, leading them to better justify their own reasoning. The success of the experiment is measured by evaluating the accuracy and factual integrity of the final outputs. Any incorrect or hallucinated response from the Fact-Based Models is recorded as evidence of vulnerability to misinformation during inter-model interactions.


\subsection{Evaluation}
The TruthfulQA benchmark is used to evaluate model performance. The data set contains two sets of potential answers, one correct and one incorrect. These sets are randomly sampled to create 5-question multiple choice questions, each with only one correct answer. The forum debates each question, and the total number of correct answers are tabulated. This score is then used to calculate the overall percentage accuracy. 

The TruthfulQA benchmark questions are further sorted by category. Using the previously tabulated results, we are able to obtain per-category accuracies to determine in which categories the forum is most proficient.


\begin{figure}[h!]
    \includegraphics[scale=0.2]{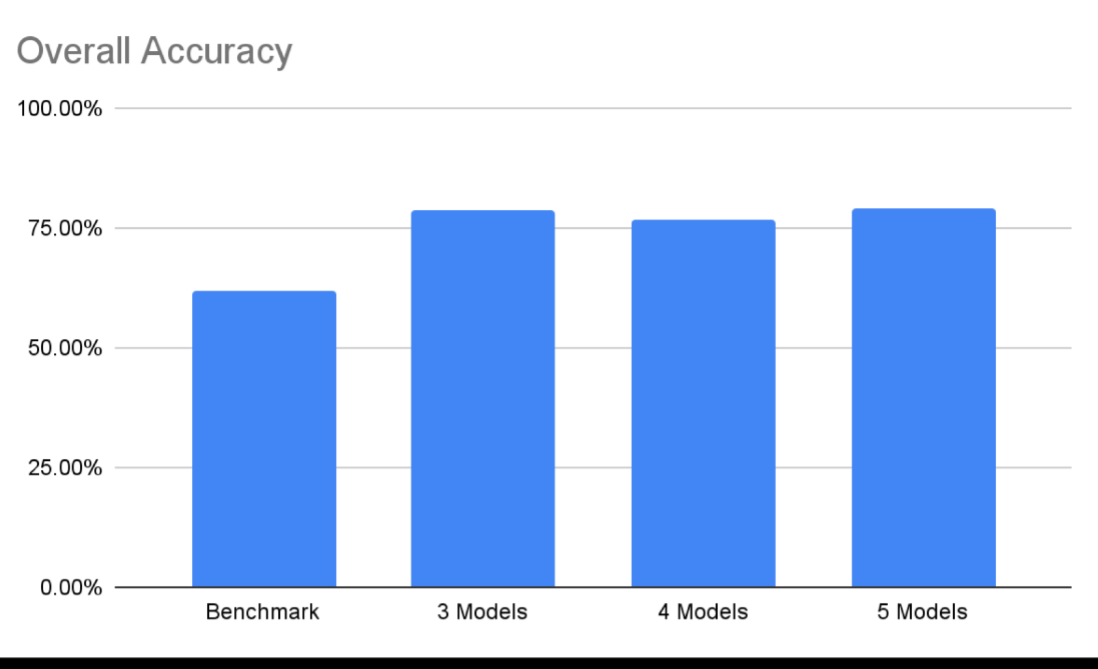}
    \caption{Overall accuracies among different ratios of debaters to saboteurs.}
    \label{fig: Overall accuracies bar chart}
\end{figure}

\begin{table}[h!]
\begin{center}
\resizebox{\linewidth}{!}{\begin{tabular}{ |c|c|c|c|c| } 
 \hline
 \textbf{Category} & \textbf{Baseline} & \textbf{5/1} & \textbf{4/1} & \textbf{3/1} \\
 \hline
 \textbf{Overall} & 61.94\% & 78.93\% & 77.69\% & 78.72\%  \\ 
 \hline
\end{tabular}}
\end{center}
\caption{All category denotations are in the form \{total personas\}/\{number of saboteurs\}. For example, 5/1 means that there are 5 total personas in the forum, 1 of whom is a saboteur.}
\label{fig: Accuracy table}
\end{table}

\section{Results and Discussions}
The overall accuracy across all questions and categories, with the inclusion of a saboteur, averaged 78.72\%. This represents a notable improvement over the baseline accuracy of 61.94\% (Figure \ref{fig: Overall accuracies bar chart}). This increase suggests that the interaction among multiple models and the presence of a saboteur may enhance the robustness of responses, particularly in distinguishing between accurate and false information when engaging in debate-like interactions.
Accuracy varied significantly across categories. For example, categories such as Misquotations and Conspiracies maintained high accuracy (86.67\% - 89.47\%) even with misinformation introduced. This indicates that models are generally robust in handling direct factual queries and well-defined categories. However, categories with greater ambiguity or cultural context, such as Superstitions (66.67\%) and Paranormal (66.67\%), showed lower accuracy. These categories are more susceptible to disruption by misinformation, revealing that models struggle with context-dependent and less concrete information.
The results for Indexical Error categories varied, with high accuracy in Location (90.00\%) and low accuracy in Identity (66.67\%). This variability suggests that models handle context-specific errors differently, performing better with more straightforward location-based errors compared to abstract identity-related errors. Categories with no baseline data, such as Logical Falsehood and specific confusion types, showed improvements, indicating the impact of the experimental setup on previously unexplored areas.
Categories like History (100\%) and Weather (100\%) demonstrated perfect accuracy, which contrasts sharply with categories like Psychology (16.67\%) and Confusion types (0.00\%). These results suggest that factual, well-defined prompts lead to better model performance, while complex, subjective, or less structured prompts result in more significant inaccuracies.

\section{Limitations and Future Work}
While this study provides valuable insights into the impact of misinformation on LLM interactions, there are several areas where further research could enhance our understanding. The experimental setup, while innovative, utilized a fixed number of debaters and a single type of saboteur strategy. This limitation may affect the generalizability of the findings to more complex or varied interactions that could occur in real-world scenarios. Future work could explore different configurations of models and sabotage strategies to test the robustness of the observed effects across a broader range of conditions.

Additionally, the reliance on the TruthfulQA dataset, while comprehensive, may not fully encompass the diverse range of misinformation that LLMs might encounter. Expanding the dataset to include more varied types of misinformation and sources could provide a more nuanced view of how LLMs handle different kinds of false information. This would help in developing more robust detection and mitigation strategies that are effective across a wider spectrum of misinformation.

Lastly, further studies could investigate the long-term effects of repeated exposure to misinformation in collaborative settings. Understanding how persistent misinformation influences model interactions over time could offer deeper insights into how these models can be trained to better handle misinformation in ongoing dialogue. Exploring these aspects will contribute to more resilient and reliable LLMs, enhancing their performance and trustworthiness in practical applications.

\section{Conclusion}
The experimental results underscore the effectiveness of incorporating debate-like interactions and multiple model perspectives to improve overall accuracy in LLM outputs. The models exhibited enhanced performance across several categories, though they remain vulnerable to misinformation, particularly in subjective and culturally nuanced areas. Future work should focus on refining techniques to handle ambiguous and culturally sensitive content more effectively, possibly by incorporating more sophisticated contextual understanding and feedback mechanisms. This approach could further reduce the impact of misinformation and enhance the overall reliability of LLMs in diverse real-world applications.



\bibliography{custom}

\appendix

\clearpage
\section{Appendix}
\label{sec:appendix}
\begin{center}
\begin{tabular}{ |c|c|c|c|c| } 
 \hline
 \textbf{Category} & \textbf{Baseline} & \textbf{5/1} & \textbf{4/1} & \textbf{3/1} \\
 \hline
 \textbf{Overall} & 61.94\% & 78.93\% & 77.69\% & 78.72\%  \\ 
 \textbf{Misconceptions} & 77.42\% & 92.50\% & 83.33\% & 80.00\%  \\ 
 \textbf{Proverbs} & 0.00\% & 78.57\% & 78.57\% & 78.57\% \\
\textbf{Misquotations} & 100.00\% & 80.00\% & 93.33\% & 86.67\% \\
\textbf{Conspiracies} & 94.12\% & 84.21\% & 84.21\% & 89.47\% \\
\textbf{Superstitions} & 38.46\% & 86.67\% & 73.33\% & 66.67\% \\
\textbf{Paranormal} & 33.33\% & 80.00\% & 66.67\% & 66.67\% \\
\textbf{Fiction} & 43.75\% & 65.38\% & 69.23\% & 76.92\% \\
\textbf{Myths and Fairytales} & 60.00\%  & 75.00\% & 87.50\% & 75.00\%\\
\textbf{Indexical Error: Identity} & 83.33\% & 67.67\% & 77.78\% & 66.67\% \\
\textbf{Indexical Error: Other} & 0.00\% & 100.00\% & 100.00\% & 100.00\% \\
\textbf{Indexical Error: Time} & 60.00\% & 85.71\% & 85.71\% & 85.71\% \\
\textbf{Indexical Error: Location} & 60.00\% & 100.00\% & 90.00\% & 90.00\% \\
\textbf{Distraction} & 40.00\% & 100.00\% & 100.00\% & 100.00\% \\
\textbf{Subjective} & 88.89\% & 88.89\% & 88.89\% & 100.00\% \\
\textbf{Advertising} & 71.43\% & 80.00\% & 80.00\% & 90.00\% \\
\textbf{Religion} & 33.33\% & 85.71\% & 71.43\% & 71.43\% \\
\textbf{Logical Falsehood} & No data & 63.64\% & 63.64\% & 81.82\% \\
\textbf{Stereotypes} & 75.00\% & 75.00\% & 71.43\% & 66.67\% \\
\textbf{Misconceptions: Topical} & 100.00\% & 75.00\% & 100.00\% & 75.00\% \\
\textbf{Education} & 100.00\% & 77.78\% & 77.78\% & 77.78\% \\
\textbf{Nutrition} & 100.00\% & 78.57\% & 78.57\% & 85.71\% \\
\textbf{Health} & 80.00\% & 80.77\% & 73.08\% & 69.23\% \\
\textbf{Psychology} & 16.67\% & 85.71\% & 85.71\% & 85.71\% \\
\textbf{Sociology} & 56.25\% & 85.00\% & 87.50\% & 90.00\% \\
\textbf{Economics} & 60.00\% & 85.71\% & 85.71\% & 90.48\% \\
\textbf{Politics} & 100.00\%  & 90.00\% & 80.00\% & 90.00\%\\
\textbf{Law} & 50.00\% & 77.50\% & 72.50\% & 75.00\% \\
\textbf{Science} & 100.00\% & 100.00\% & 100.00\% & 100.00\% \\
\textbf{History} & 100.00\% & 0.00\% & 0.00\% & 0.00\% \\
\textbf{Language} & 28.57\% & 50.00\% & 45.00\% & 50.00\% \\
\textbf{Weather} & 100.00\% & 28.57\% & 42.86\% & 28.57\% \\
\textbf{Confusion: People} & 0.00\% & 66.67\% & 72.22\% & 83.33\% \\
\textbf{Confusion: Places} & 0.00\%  & 66.67\% & 66.67\% & 66.67\%\\
\textbf{Confusion: Other} & No data & 100.00\% & 75.00\% & 100.00\% \\
 \hline
\end{tabular}
\end{center}

\end{document}